\documentclass[runningheads]{llncs}
\usepackage{graphicx}
\usepackage{ltablex}
\usepackage{booktabs} 
\usepackage{amsmath}
\usepackage[table]{xcolor}
\usepackage{pifont}
\usepackage{multirow}
\usepackage{xspace}
\usepackage{bold-extra}
\usepackage[misc]{ifsym}

\newcommand{\model}[1]{\textsc{#1}\xspace}
\newcommand{\moqa}{\model{moqa}}
\newcommand{\filter}{\model{fltr}}
\newcommand{\bert}{\model{bertqa}}

\newcommand{\ndcga}{$\mathcal{N}_{A}$\xspace}
\newcommand{\ndcgu}{$\mathcal{N}_{U}$\xspace}
\newcommand{\ndcgau}{$\mathcal{N}_{A+U}$\xspace}

\newcommand{\tuning}{\model{thrs}}
\newcommand{\moqatuning}{\moqa{}$+$\tuning}
\newcommand{\berttuning}{\bert{}$+$\tuning}
\newcommand{\filtertuning}{\filter{}$+$\tuning}

\newcommand{\conform}{\model{imcp}}
\newcommand{\moqacp}{\moqa{}$+$\conform}
\newcommand{\bertcp}{\bert{}$+$\conform}
\newcommand{\filtercp}{\filter{}$+$\conform}

\newcommand{\secref}[2][]{Section#1~\ref{sec:#2}}

\newcommand{\tabref}[2][]{Table#1~\ref{table:#2}}
\newcommand{\figref}[2][]{Figure#1~\ref{fig:#2}}

\newcommand{\eqnref}[2][]{Equation#1~(\ref{eqn:#2})}

\usepackage{xcolor}

\begin{document}
    \title{Less is More: Rejecting Unreliable Reviews for
Product Question Answering}
	%
	%
	\author{\small Shiwei Zhang\inst{1,3} \and
	    Xiuzhen Zhang\inst{1}\textsuperscript{\Letter}  \and
	    Jey Han Lau\inst{2}  \and
	    Jeffrey Chan\inst{1}  \and
	    Cecile Paris\inst{3}}
	\institute{RMIT University, Australia 
		\email{\{firstname.lastname\}@rmit.edu.au}\\
		\and
	   The University of Melbourne, Australia
	  \email{jeyhan.lau@gmail.com}\\
	   \and
	   CSIRO Data61, Australia
	  \email{cecile.paris@data61.csiro.au}}

	%
    \authorrunning{S. Zhang et al.}
	%
	%
    \maketitle              
    \begin{abstract}
        Promptly and accurately answering questions on products  is 
important for e-commerce applications.
        Manually answering product questions  (e.g. on community 
question answering platforms) results in slow response and does not 
scale.
        Recent studies show that product reviews are a good source for 
real-time, automatic product question answering (PQA). In the literature, 
PQA is formulated  as a retrieval problem with the goal to search for 
the most relevant reviews to answer a given product question. 
        In this paper, we focus on the issue of \textit{answerability} and \textit{answer reliability} for PQA using 
reviews.
        Our investigation is based on the intuition that many questions 
may not be answerable with a finite set of reviews.
       When a question is not answerable, a system should return nil
answers rather than providing a list of irrelevant reviews, which can 
have significant negative impact on user experience.
Moreover, for answerable questions, only the most relevant 
reviews that answer the question should be included in the result. 
       We propose a conformal prediction based framework to improve the 
reliability of PQA systems, where we reject unreliable answers so that 
the returned results are  more concise and accurate at answering the 
product question, including returning nil answers for unanswerable 
questions.
        Experiments on a widely used Amazon dataset show encouraging results 
        of our proposed framework. More broadly, our results demonstrate 
a novel and effective application of conformal methods to a retrieval 
task.
		
		\keywords{Product Question Answering  \and Unanswerable Questions \and Conformal Prediction.}
	\end{abstract}
	
	\section{Introduction}
	On e-commerce websites such as 
	Amazon\footnote{\url{https://www.amazon.com/}.} and 
	TaoBao\footnote{\url{https://world.taobao.com/}.}, customers often ask 
	product-specific questions prior to a purchase.
    With the number of product-related questions (queries) growing, efforts to 
    answer these queries manually in real time is increasingly 
infeasible.
	Recent studies found that product reviews are a good source to 
	extract helpful information as 
	answers~\cite{mcauley2016addressing,zhao2019riker,zhang2019icdm}.
	
	To illustrate this, in Table~\ref{table:casestudy} Q1 poses a question 
	about the purpose of the chain on the side of a  grill, and the first 
	review addresses the question.
	The core idea of state-of-the-art PQA models is to take advantage of 
	existing product reviews and find relevant reviews that answer questions 
	automatically.
	In general, most PQA models implement a relevance function to rank 
	existing reviews based on how they relate to a question.
    Some directly present a fixed number (typically 10) of the top 
    ranked reviews as answers for the 
question~\cite{mcauley2016addressing,zhao2019riker,zhang2019icdm}, while 
others generate natural-language answers based on the relevant reviews 
~\cite{gao2019product,chen2019driven}.
	
	However, not every question can be answered by reviews:
    the existing set of reviews may not contain any relevant answers for 
the question, or a question may be poorly phrased and difficult to 
interpret and therefore requires additional clarification.
	Q2 in Table~\ref{table:casestudy} is an example of an unanswerable 
	question. The user wants to know whether the notebook comes with Dell's 
	warranty, but none of the reviews discuss anything about warranty.
	In such a case, a system should abstain from returning any reviews and 
    forward the question to the product seller. In the PQA literature, the 
issue of answerability is largely unexplored, and as such evaluation 
focuses on simple ranking performance without penalising systems that 
return irrelevant reviews. 
	
	\begin{table}[t]
        \begin{center}	\caption{Example of an answerable (Q1) and an 
                unanswerable question (Q2). Green denotes high 
probability/confidence scores, and red otherwise.}
			\begin{tabular}{ p{1cm}p{10cm}@{\;\;\;}}
                \toprule
				\multicolumn{2}{l}{Q1: What is the chain for on the side?} \\
				\raisebox{-.5\totalheight}{\includegraphics[width=0.08\textwidth, height=8mm]{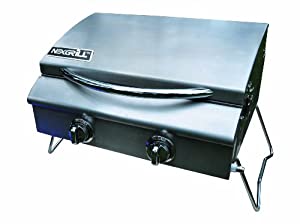}}
				& 
				\begin{tabular}{p{7.4cm}@{\;\;\;}ccc} 
					Top 3 Ranked Reviews & Prob & Conf & Accept \\ \midrule
					- This was driving me crazy but i see that another reviewer explained that grill has wire clip on chain to be used as extended match holder for igniting the gas if the spark mechanism fails to work or is worn out as sometimes happens with any gas grill.  & \cellcolor{green!25}0.99  & \cellcolor{green!25}0.82 &\multicolumn{1}{c}{ \ding{52}}\\ 
					- PS Could not figure out the purpose of that little chain with the clip attached to the outside of the grill - even after reading entire manual. & \cellcolor{green!25}0.95 & \cellcolor{red!25}0.54 & \multicolumn{1}{c}{ \ding{56}}\\ 
					- It is to replace an old portable that I have been using for about 10 years.' & \cellcolor{green!25}0.91 & \cellcolor{red!25}0.40 & \multicolumn{1}{c}{ \ding{56}}\\ 
				\end{tabular}		
				\\ \midrule 
				
				\multicolumn{2}{l}{Q2: Does this Dell Inspiron 14R i14RMT-7475s come with dell's warranty?} \\
				\raisebox{-.5\totalheight}{\includegraphics[width=0.08\textwidth, height=8mm]{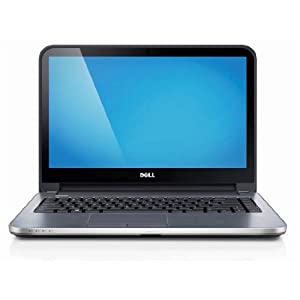}}
				& 
				\begin{tabular}{ p{7.4cm}@{\;\;\;}ccc} 
					Top 3 Ranked Reviews & Prob & Conf & Accept \\ \midrule
					- I don't really recommend the PC for people who wants install heavy games programs. & \cellcolor{green!25}0.74 & \cellcolor{red!25}0.48 & \multicolumn{1}{c}{ \ding{56}}\\ 
					- The computer is nice, fast, light, ok. & \cellcolor{red!25}0.12 &\cellcolor{red!25}0.01 & \multicolumn{1}{c}{ \ding{56}}\\ 
					- I bought the computer for my daughter. &\cellcolor{red!25} 0.05 &\cellcolor{red!25}0.00 & \multicolumn{1}{c}{ \ding{56}}\\ \end{tabular}		
				\\ \bottomrule 
			\end{tabular}
			\label{table:casestudy}
		\end{center}
		
	\end{table}
	
	
	That said, the question answerability issue has begun to draw some attention in 
	machine comprehension (MC).
    Traditionally, MC models assume the correct answer span always exists in the context passage
for a given question.
	As such, these systems will give an incorrect (and often embarrassing) 
	answer when the question is not answerable. This motivates the 
	development of better comprehension systems that can distinguish between 
	answerable and unanswerable questions.
	Since SQuAD --- a popular MC dataset --- released its second 
	version~\cite{raj2018unanswerabe} which contains approximately 50,000 
	unanswerable questions, various MC models have been proposed to detect 
	question-answerability in addition to predicting answer 
	span~\cite{hu2019unanswerable,sun2018unanswerable,godin2019unanswerable,huang2019unanswerable}.
	The MC models are trained to first detect whether a 
	question is answerable or unanswerable and then find answers 
	to answerable questions.
	 \cite{sun2019webqa} proposed a risk controlling framework to increase 
    the reliability of MC models, where risks are quantified based on 
the extent of incorrect predictions, for both answerable and unanswerable 
questions.
    Different from MC which always has one answer, PQA is a ranking 
problem where there can be a number of relevant reviews/answers. As 
such, PQA is more challenging, and
risk models designed for MC cannot be trivially adapted for PQA.


    In this paper,  we focus on the problem of answer reliability for 
    PQA.  Answer reliability is generalisation of answerability.  
Answerability is a binary classification problem where answerable 
questions have only one answer (e.g.\ MC).  In our problem setting, a 
product question can have a variable number of reliable answers 
(reviews), and   
    questions with nil reliable answers are the unanswerable questions.	
    The challenge is therefore on how we can estimate the reliability of 
answers.
    As our paper shows,  the naive approach of thresholding based on the 
predicted probability from PQA models
is not effective for estimating the reliability of candidate answers.

	
    We tackle answer reliability by introducing a novel application of 
    conformal predictors as a rejection model.
    The rejection model~\cite{herbei2006rejection} is a technique proposed 
    to reduce the misclassification rate for risk-sensitive classification applications.
    The risk-sensitive prediction framework consists of two models: a classifier that outputs 
    class probabilities given an example, and a rejection model that 
measures the confidence of its prediction and rejects unconfident 
prediction.
	In our case, given a product question, the PQA model makes a 
	prediction on the relevance for each review, and the rejection model 
	judges the reliability for each prediction and returns only reviews that 
	are relevant \textit{and} reliable as answers.
    As an example, although the positive class probabilities given by 
    the PQA model to the top 3 reviews are very high in Q1 
(\tabref{casestudy}), the rejection model would reject the last two 
because their confidence scores are low.
	Similarly for Q2, even though the first review has high relevance,
    the rejection model will reject all reviews based on the confidence 
scores, and return an empty list to indicate that this is an 
unanswerable question.
	
	
	For the PQA models, we explore both classical machine learning 
	models~\cite{mcauley2016addressing} and BERT-based neural 
	models~\cite{zhang2019icdm}. For the rejection model, we use an 
	Inductive Mondrian Conformal Predictor 
	(IMCP)~\cite{gamm2019conformal,vovk2005conformal,shafer2008conformal,tocca2019conformal}.  
	IMCP is a popular non-parametric conformal predictor used in a range of 
	domains, from drug discovery~\cite{carl2017conformal} to chemical 
	compound activity prediction~\cite{tocca2019conformal}.
    The challenge of applying a rejection model to a PQA model is how 
we define the associated risks in PQA. Conventionally, a rejection 
mode is adopted to reduce the misclassification rate, but for PQA 
we need to reduce the error of including irrelevant results.
    We propose to use IMCP to transform review relevance scores (or 
    probabilities) into confidence scores and tune a threshold to 
minimize risk (defined by an alternative ranking metric that penalises 
inclusion of irrelevant results), and reject reviews whose confidence 
scores are lower than the threshold.
	
    To evaluate the effectiveness of our framework in producing relevant 
    and reliable answers, we conduct a crowdsourcing study using 
{Mechanical Turk}\footnote{\url{https://www.mturk.com/}.} to acquire the 
relevance of reviews for 200 product-specific questions on 
Amazon~\cite{mcauley2016addressing}. We found encouraging results, 
demonstrating the applicability of rejection models for PQA to handle 
unanswerable questions.  
To facilitate replication and future research, 
we release source code and data used in our 
experiments.\footnote{\url{https://github.com/zswvivi/ecml\_pqa}}

	\section{Related Work}
	In this section, we survey three related topics to our work: product 
	question answering, question answerability and conformal predictors.
	
	\subsection{Product Question Answering}
	Existing studies on product-related question answering using reviews can 
	be broadly divided into extractive 
	approaches~\cite{mcauley2016addressing,zhang2019icdm} and generative 
	approaches~\cite{gao2019product,chen2019driven}.
    For extractive approaches, relevant reviews or review snippets are 
    extracted from reviews to answer questions, while for the generative 
approaches natural answers are further generated based on the review 
snippets. In both approaches, the critical step is to first identify 
relevant reviews that can answer a given question.
	
    The key challenge in PQA is the lack of ground truth, i.e.\ there is 
limited data with  annotated relevance scores between questions and 
reviews.
	Even with crowdsourcing, the annotation work is prohibitively expensive, 
	as a product question may have a large number of reviews; and more so if 
	we were to annotate at sentence level (i.e.\ annotating whether a 
	sentence in a review is relevant to a query), which is typically the 
	level of granularity that PQA studies work 
	with~\cite{mcauley2016addressing,zhang2019icdm}.
    For that reason, most methods adopt a distant supervision approach that 
    uses existing question and answer pairs from an external source (e.g. the community question answer platform)  
    as supervision.
    The first of such study is the Mixture of Opinions for Question 
    Answering (\moqa) model proposed by~\cite{mcauley2016addressing}, which is 
inspired by the mixture-of-experts classifier~\cite{jacobs1991moe}.  
Using answer prediction as its objective, \moqa decomposes the task into 
learning two relationships: (1) relevance between a question and a 
review; and (2) relevance between a review and an answer, sidestepping 
the need for ground truth relevance between a question and a review. 
    \cite{zhang2019icdm} extends \moqa by parameterising the relevance 
    scoring functions using BERT-based models~\cite{devlin2019bert} and 
found improved performance.
	
	Another recent work~\cite{zhao2019riker} learns deep
	representations of words between existing questions and answers.
	To expand the keywords of a query, the query words are first mapped to 
	their continuous representations and similar words in the latent space 
	are included in the expanded keywords.
	To find the most relevant reviews, the authors use a standard 
	keyword-based retrieval method with the expanded query and found 
	promising results.
	
	\subsection{Unanswerable Questions}

	There are few studies that tackle unanswerable questions in PQA. One 
    exception is \cite{gupta2019ijcai}, where they develop a new PQA 
dataset with labelled unanswerable questions. That said, the author 
frame PQA as a classification problem, where the goal is to find an 
answer span in top review snippets (retrieved by a search engine), and 
as such the task is more closely related to machine comprehension.
	
	
	In the space of MC, question answerability drew some attention when 
    SQuAD 2.0~\cite{raj2018unanswerabe} was released, which includes 
over 50,000 unanswerable questions created by crowdworkers. 
	Several deep learning MC systems have since been proposed to tackle 
	these unanswerable questions.  \cite{hu2019unanswerable} proposed a 
	read-then-verify system with two auxiliary losses, where the system 
	detects whether a question is answerable and then checks the validity of 
	extracted answers.
	\cite{sun2018unanswerable} proposed a multi-task learning model that 
	consists of three components: answer span prediction, question 
	answerability detection and answer verification.
	
	More generally in question answering (QA), Web QA is an open-domain 
	problem that leverages Web resources to answer questions, e.g.\ 
	TriviaQA~\cite{joshi2017triviaqa} and SearchQA~\cite{dunn2017searchqa}.
    \cite{sun2019webqa} introduced a risk control framework to manage the 
    uncertainty of deep learning models in Web QA. The authors argue 
that there are two forms of risks, by returning: (1) wrong answers for 
answerable questions; and (2) any answers for unanswerable questions.  
The overall idea of their work is similar to ours, although their 
approach uses a probing method that involves intermediate layer outputs 
from a neural model, which is not applicable to non-neural models such 
as \moqa.
	
	\subsection{Conformal Predictors}
	
    To measure the reliability of prediction for an unseen example, 
conformal predictors (CP) compares how well the unseen example 
\textit{conforms} to previously seen examples.
	Given an error probability $\epsilon$, CP is guaranteed to produce a 
	prediction region with probability $1-\epsilon$ of containing the true 
	label $y$, thereby offering a means to control the error rate.
	CP has been applied to many different areas, from drug 
	discovery~\cite{sun2017conformal,cortes2019conformal,tocca2019conformal} 
	to image and text classification~\cite{card2019conformal}.
    \cite{cortes2019conformal} proposed a neural framework using Monte 
    Carlo Dropout~\cite{gal2016dropout} and CP to compute reliable 
errors in prediction to guide the selection of active molecules in 
retrospective virtual screen experiments.
    In~\cite{card2019conformal}, the authors replace the softmax layer 
    with CP
	to predict labels based on a weighted sum of training instances for 
	image and text classification.

\section{Methodology}

\begin{figure}
	\centering
	\includegraphics[width=0.8\linewidth]{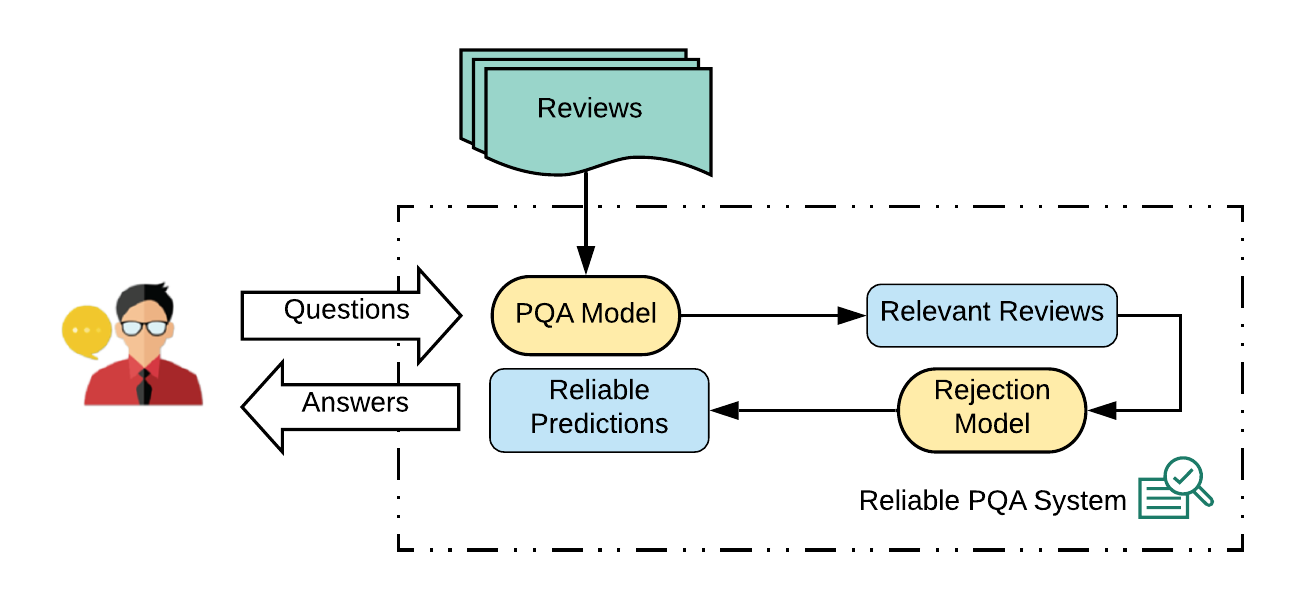}
    \caption{Proposed PQA $+$ a rejection model framework.}
	\label{fig:framework}
\end{figure}

Our proposed framework consists of two components: a PQA model that 
predicts the relevance of reviews given a product query, and a rejection 
model that rejects unconfident reviews and produce only confident 
reviews as answers. An illustration of our framework is presented in 
\figref{framework}.

More specifically, the PQA component (a binary classifier) models the 
function $\hat{y} = F(r, q)$ where $r$ is a review, $q$ is a product 
question, and $\hat{y}$ is the probability of the positive class, i.e.  
the review is relevant to the question.

The rejection model takes the output ${\hat{y}}$ from the PQA model and 
transforms it into a confidence score ${y}^{*}$.  Given the confidence 
score, we can then set a significance level $\epsilon \in [0,1]$ to 
reject/accept predictions.  E.g.\ if the confidence score of the 
positive class (review is relevant to the question) is 0.6 and $\epsilon 
$ is 0.5, then we would accept the positive class prediction and return 
the review as relevant.  On the other hand, if $\epsilon $ is 
0.7, we would reject the review.


\subsection{PQA Models}

We explore 3 state-of-the-art PQA models for our task:

\subsubsection{\moqa}
\cite{mcauley2016addressing} is inspired by the mixture-of-experts 
classifier~\cite{jacobs1991moe}. In a mixture of experts classifier, a 
prediction is made by a weighted combination of a number of weak 
classifiers.
Using answer prediction ($P(a|q)$) as its objective, \moqa decomposes 
the problem into: $P(a|q) = \sum_{r} P(a|r) P(r|q)$, where $a, q, r =$ 
answer, query, review respectively. Each review can be interpreted as an 
``expert'', where it makes a prediction on the answer ($P(a|r)$) and its 
prediction is weighted by its confidence ($P(r|q)$). The advantage of 
doing this decomposition is that the learning objective is now $P(a|q)$, 
which allows the model to make use of the abundance of existing 
questions and answers on e-commerce platforms. In practice, \moqa is 
optimised with a maximum margin objective: given a query, the goal is to 
score a real answer higher than a randomly selected non-answer. To model 
the two relevance functions ($P(a|r)$ and $P(r|q)$), \moqa uses 
off-the-shelf pairwise similarity function such as BM25+ and ROUGE-L and 
a learned bilinear scoring function that uses bag-of-words 
representation as input features. After the model is trained, the 
function of interest is $P(r|q)$,  which can be used to score a review 
for a given query.
We use the open source implementation and its default optimal 
configuration.\footnote{\url{https://cseweb.ucsd.edu/$\sim$jmcauley}}


\subsubsection{\filter}
\cite{zhang2019icdm} is a BERT classifier for answer prediction. Using 
existing question and answer pairs on e-commerce platforms, 
\cite{zhang2019icdm} fine-tunes a pre-trained BERT to classify answers 
given a question. After fine-tuning, \filter is used to classify 
\textit{reviews} for a given question. \filter can be seen as a form of 
zero-shot domain transfer, where it is trained in the (answer, question) 
domain but at test time it is applied to the (review, question) domain.
We use the open-source implementation and its default optimal 
configuration.\footnote{\url{https://github.com/zswvivi/icdm\_pqa}}

\subsubsection{\bert} \cite{zhang2019icdm} is an extension of \moqa, 
which uses the same mixture of experts framework, but parameterises the 
relevance functions ($P(a|r)$ and $P(r|q)$) with neural networks: BERT.
\bert addresses the vocabulary/language mismatch between different 
domains (e.g. answer vs. review, or review vs. query) using 
contextualised representations and fine-grained comparison between words 
via attention. \cite{zhang2019icdm} demonstrates that this neural 
parameterisation substantially improves review discovery compared to 
\moqa.
The downside of \bert is its computational cost: while \moqa can be used 
to compute the relevance for every review for a given query (which can 
number from hundreds to thousands), this is impractical with \bert. To 
ameliorate this, \cite{zhang2019icdm} propose using \filter to 
pre-filter reviews to reduce the set of reviews to be ranked by  
\bert.\footnote{The original implementation uses a softmax activation 
function to compute $P(r|q)$ (and so the probability of all reviews sum 
up to one); we make a minor modification to the softmax function and use 
a sigmoid function instead (and so each review produces a valid 
probability distribution over the positive and negative classes).}


\subsection{Rejection Model}

In a classification task, a model could make a wrong prediction for a 
difficult instance, particularly when the positive class probability is 
around 0.5 in a binary task. For medical applications such as tumor 
diagnostic, misclassification can have serious consequences. In these 
circumstances, rejection techniques are used to reduce misclassification 
rate by rejecting unconfident or unreliable 
predictions~\cite{herbei2006rejection}.

To apply rejection techniques to PQA, we need to first understand the 
associated risks in PQA.  There are forms of risks in PQA: (1) including 
irrelevant reviews as part of its returned results for an answerable 
question; and (2) returning any reviews for an unanswerable question.  
This is similar to the risks proposed for Web QA~\cite{sun2019webqa}, 
although a crucial difference is that PQA is a ranking problem (output 
is a list of reviews). The implication is that when deciding the 
$\epsilon$ threshold to guarantee a ``error rate'', we are using a 
ranking metric as opposed to a classification metric. As standard 
ranking metric such as normalised discounted cumulative gain (NDCG) is unable 
to account for unanswerable questions, we explore alternative metrics 
(detailed in \secref{evaluation-metric}).


We propose to use conformal predictors 
(CP)~\cite{vovk2005conformal,shafer2008conformal} as the rejection 
model. Intuitively, for a test example, CP computes a nonconformity 
score that measures how well the new example \textit{conforms} to the 
previously seen examples as a way to estimate the reliability of the 
prediction.
CP is typically applied to the output of other machine learning models, and 
can be used in both classification and regression tasks.
There are two forms of CP, namely Inductive CP (ICP) and Transductive CP 
(TCP).
The difference between them is their training scheme: in TCP, the 
machine learning model is updated for each new examples (and so requires 
re-training) to compute the nonconformity score, while in ICP the model 
is trained once using a subset of the training data, and the other 
subset --- the \textit{calibration set} --- is set aside to be used to 
compute nonconformity scores for a new example.
As such, the associated computation cost for the transductive variant is 
much higher due to the re-training.
We use the {Inductive Mondrian Conformal Predictor (IMCP)} for our 
rejection model, which is a modified ICP that is better at handling 
imbalanced data.  When calculating a confidence score, IMCP additionally 
conditions it on the class label. In PQA, there are typically a lot more 
irrelevant reviews than relevant reviews for given a query, and so IMCP 
is more appropriate for our task.



\begin{table}[t]
\begin{center}
\caption{An example of predicted labels with different choices of 
significance level. $p$-value for the positive (1) and negative (0) 
label is 0.65 and 0.45 respectively.}
\label{table:prediction_region}
\begin{tabular}{c@{\;\;\;\;}c@{\;\;\;\;}c}
\toprule
$\mathbf{\epsilon}$ & \textbf{Predicted Labels} & \textbf{Outcome} \\
\midrule
0.05 & \{0, 1\} & Rejected \\
0.45 & \{1\} & Accepted \\
0.75 & \{\} & Rejected \\
\bottomrule
\end{tabular}
\end{center}
\end{table}

Given a bag of calibration examples $\{ (x_1,y_1),...(x_n,y_n)\}$ and a 
binary classifier ${\hat{y}} = f(x)$, where $n$ is number of calibration 
examples, $x$ the input, $y$ the true label and ${\hat{y}}$ the 
predicted probability for the positive label, we compute a nonconformity 
score, $\alpha(x_{n+1})$, for a new example $x_{n+1}$ using its inverse 
probability:
\begin{equation*}
\alpha(x_{n+1}) = -f(x_{n+1})
\end{equation*}

As IMCP conditions the nonconformity score on the label, there are 
2 nonconformity scores for $x_{n+1}$, one for the positive label, and 
  one for the negative label:
\begin{align*}
\alpha(x_{n+1}, 1) &= -f(x_{n+1}) \\
\alpha(x_{n+1}, 0) &= - (1-f(x_{n+1}))
\end{align*}


We then compute the confidence score ($p$-value) for $x_{n+1}$ 
conditioned on label $k$ as follows:
\begin{equation}
        p(x_{n+1},k)= \frac{\sum_{i=0}^{n} {I}(\alpha(x_i,k) \geq \alpha 
(x_{n+1},k))}{n+1}
\label{eqn:pvalue}
\end{equation}
where $I$ is the indicator function.

Intuitively, we can interpret the $p$-value as a measure of how 
confident/reliable the prediction is for the new example by comparing 
its predicted label probability to that of the calibration examples.


Given the $p$-values (for both positive and negative labels), the 
rejection model accepts a prediction for all labels $k$ where 
$p(x_{n+1},k) > \epsilon$, where $\epsilon \in [0,1]$ is the 
{significance level}. We present an output of the rejection model in 
\tabref{prediction_region} with varying $\epsilon$, for an example whose  
$p$-values for the positive and negative labels are 0.65 and 0.45 
respectively.  Depending on  $\epsilon$, the number of predicted labels 
ranges from zero to 2. For PQA, as it wouldn't make sense to have both 
positive and negative labels for an example (indicating a review is both 
relevant and irrelevant for a query), we reject such an example and 
consider it an unreliable prediction.

	

\section{Experiments}
\subsection{Data}

We use the Amazon dataset developed by~\cite{mcauley2016addressing} for 
our experiments. The dataset contains QA pairs and reviews for each 
product.
We train \moqa, \bert and \filter on this data, noting that \moqa and 
\bert leverages both QA pairs and reviews, while \filter uses only the 
QA pairs. After the models are trained, we can use the $P(r|q)$ 
relevance function to score reviews given a product question.

To assess the quality of the reviews returned for a question by the PQA 
models, we ask crowdworkers on Mechanical Turk to judge how well a 
review answers a question.
We randomly select 200 questions from four categories (``Tools and Home 
Improvement'',``Patio Lawn and Garden'',``Baby'' and ``Electronics''), 
and
pool together the top 
10 reviews returned by the 3 PQA models (\moqa, \bert and 
   \filter),\footnote{Following the original papers, a ``review'' is 
technically a ``review sentence'' rather than the full review.} 
resulting in approximately 
20 to 
30 reviews per question (total number of reviews $=$ 4,691).\footnote{To 
   control for quality, we insert a control question with a known answer 
(from the QA pair) in every 3 questions. Workers who consistently give 
low scores to these control questions are filtered out.}


\begin{table}[t]
    \centering  \caption{Answerable question statistics.}
	\label{table:annotation}
    \begin{tabular}{l@{\;\;\;\;}r@{\;\;\;}r@{\;\;\;}r@{\;\;\;}r@{\;\;\;}r} \toprule
        \textbf{Relevance Threshold} & 2.00 & 2.25 & 2.50& 2.75 & 3.00 
        \\ \midrule
        \textbf{\#Relevant Reviews}   & 640 & 351 & 175 & 71 & 71 \\
        \textbf{\#Answerable Questions}  & 170 & 134 & 89 & 44 & 44 \\
        \textbf{\%Answerable Questions} & 85\% & 67\% & 45\% & 22\% & 
        22\% \\
		\bottomrule
	\end{tabular}
\end{table}

In the survey, workers are presented with a pair of question and review, 
 and they are asked to judge how well the review answers the question on 
an ordinal scale:  0 (completely irrelevant), 
1 (related but does not answer the question), 2 (somewhat answers the 
  question) and  3 (directly answers the question). Each question/review 
pair is annotated by 3 workers, and the final relevance score for each 
review is computed by taking the mean of 3 scores.

Given the annotated data with relevance scores, we can set a relevance 
threshold to define a cut-off when a review answers a question, allowing 
us to control how precise we want the system to be (i.e.\ a higher 
threshold implies a more precise system).
We present some statistics in \tabref{annotation} with different 
relevance thresholds.  For example, when the threshold is set to 2.00, 
it means a review with a relevance score less than 2.00 is now 
considered irrelevant (and so its score will be set to 0.00), and a 
question where all reviews have a relevance score less than 2.00 is now 
unanswerable.
The varying relevance thresholds will produce a different distribution 
 of relevant/irrelevant reviews and answerable/unanswerable questions; 
at the highest threshold (3.00), only a small proportion of the reviews 
are relevant (but they are all high-quality answers), and most questions 
are unanswerable.

\subsection{Evaluation Metric}
\label{sec:evaluation-metric}

\begin{table}[t]
	\centering  \caption{NDCG' examples.}
	\label{table:ndcg}
    \begin{tabular}{c@{\;\;\;}l@{\;\;\;}l@{\;\;\;}l} \toprule
        \textbf{Question Type} & \textbf{Systems} & \textbf{Doc
        List} & \textbf{NDCG'}  \\ \midrule
		\multirow{3}{*}{Answerable} 
		& System A  & 111   & 1.000 \\
		& System B  & 11100 & 0.971 \\
		&System C  & 11    & 0.922 \\  \midrule
		\multirow{3}{*}{Unanswerable}
        &System A  & $\emptyset$   & 1.000 \\
		&System B  & 00 & 0.500 \\
		&System C  & 000 & 0.431 \\  
		\bottomrule
	\end{tabular}	
\end{table}

As PQA is a retrieval task, it is typically evaluated using ranking 
metrics such as normalised discounted cumulative gain 
(NDCG)~\cite{jarvein2002ndcg}.  Note, however, that NDCG is not designed 
to handle unanswerable queries (i.e. queries with no relevant 
documents), and as such isn't directly applicable to our task.
We explore a variant, NDCG', that is designed to work with unanswerable 
queries \cite{liu2016ndcg}. The idea of NDCG' is to ``quit while 
ahead'': the returned document list should be truncated earlier rather 
than later, as documents further in the list are more likely to be 
irrelevant.
To illustrate this, we present an answerable question in \tabref{ndcg}.  
Assuming it has three relevant documents (1 represents a relevant 
document and 0 an irrelevant document), System A receives a perfect 
NDCG' score while System B is penalised for including 2 irrelevant 
documents. System C has the lowest NDCG' score as it misses one relevant 
document.
The second example presents an unanswerable question.  The ideal result 
is the empty list ($\emptyset$) returned by System A, which receives a 
perfect score. Comparing System B to C, NDCG' penalises C for including 
one more irrelevant document.

NDCG' appends a terminal document (\textit{t}) to the end of the 
document list returned by a ranking system.
For example, ``111'' $\rightarrow$ ``111\textit{t}'', and ``11100'' 
$\rightarrow$ ``11100\textit{t}''.
The corresponding gain value for the terminal document \textit{t} is  
$r_t$, calculated as follows:
\begin{equation*}
r_t = \begin{cases} 
1 &\textbf{if \(R=0\)} \\
\sum_{i=1}^{d}r_i/R &\textbf{if \(R>0\)}
\end{cases}     \end{equation*}
where $R$ is the total number of ground truth relevant documents, and 
$r_i$ is the relevance of document $i$ in the list. As an  example, for 
the document list ``11'' produced by System C, $r_t = \frac{1}{3} + 
\frac{1}{3} = \frac{2}{3}$.

Given $r_t$, we compute NDCG' for a ranked list of $d$ items as follows:
\begin{equation*}
\text{NDCG'}_d =  \frac{\text{DCG}_{d+1}\langle r_1,r_2,...,r_d,r_t 
\rangle}{\text{IDCG}_{d+1}}
\end{equation*}

With NDCG, for an unanswerable question like the second example, both 
System B (``00'') and System C (``000'') will produce a score of zero, 
and so it fails to indicate that B is technically better. NDCG' solves 
this problem by introducing the terminal document score $r_t$.

In practice, the relevance of our reviews is not binary (unlike the toy 
examples). That is, the relevance of each review is a mean relevance 
score from 3 annotators, and ranges from 0--3. Note, however, that given 
a particular relevance threshold (e.g.\ 2.0), we mark all reviews under 
the threshold as irrelevant by setting their relevance score to 0.0.


In our experiments, we compute NDCG' up to a list of 10 reviews ($d =$ 
10), and separately for answerable (\ndcga) and unanswerable questions 
   (\ndcgu). To aggregate NDCG'  over two question types (\ndcgau), we 
compute the geometric mean:
\begin{equation*}
    \text{\ndcgau} = \sqrt{\text{\ndcga} \times \text{\ndcgu}}
\end{equation*}

We use geometric mean here because we want an evaluation metric that 
favours a balanced performance between answerable and unanswerable 
questions~\cite{kubat1998images}.  In preliminary experiments, we found 
that micro-average measures will result in selecting a system that 
always returns no results when a high relevance threshold is selected 
(e.g.\ $\geq$ 
2.50 in \tabref{annotation}) since a large number of questions are 
  unanswerable. This is undesirable in a real application where choosing 
a high relevance threshold means we want a very precise system, and not 
one that never gives any answers.


\subsection{Experimented Methods}

We compare the following methods in our experiments:

\textbf{Vanilla PQA Model}: a baseline where we use the top-10 reviews 
returned by a PQA model (\moqa, \filter or \bert) without any 
filtering/rejection.
	
\textbf{PQA Model$+$\tuning}: a second baseline where we tune a 
threshold based on the review score returned by a PQA model (\moqa, 
\filter or \bert) to truncate the document list. We use leave-one-out 
cross-validation for tuning. That is, we split the 200 annotated 
questions into 199 validation examples and 1 test example, and find an 
optimal threshold for the 199 validation examples based on \ndcgau.  
Given the optimal threshold, we then compute the final \ndcgau on the 1 
test example. We repeat this 200 times to get the average \ndcgau 
performance.

	
\textbf{PQA Model$+$\conform}: our proposed framework that combines PQA 
    and \conform as the rejection model. Given a review score by a PQA 
model, we first convert the score into probabilities,\footnote{This step 
is only needed for \moqa, as \bert and \filter produce probabilities in 
the first place.  For \moqa, we convert the review score into a 
probability applying a sigmoid function to the log score.} and then 
compute the $p$-values for both positive and negative labels 
(\eqnref{pvalue}). We then tune the significance level $\epsilon$ to 
truncate the document list, using leave-one-out cross-validation as 
before. As IMCP requires calibration data to compute the $p$-value, the 
process is a little more involving.  We first split the 200 questions 
into 199 validation examples and 1 test examples as before, and within 
the 199 validation examples, we do another leave-out-out: we split them 
into 198 calibration examples and 1 validation example, and compute the 
$p$-value for the validation example based on the 198 calibration 
examples. We then tune $\epsilon$ and find the optimal threshold that 
gives the best \ndcgau performance on the single validation performance, 
and repeat this 199 times to find the best overall $\epsilon$. Given 
this $\epsilon$, we then compute the \ndcgau performance on the 1 test 
example. This whole process is then repeated  200 times to compute the 
average \ndcgau test performance.


\begin{table}[t!]
	\centering  \caption{Model performance; boldface indicates optimal \ndcgau performance for a PQA model.}
	\label{table:results}
	\begin{tabular}{c@{\;\;\;\;}l@{\;\;\;\;}c@{\;\;\;\;}c@{\;\;\;\;}c} \toprule
		\textbf{Relevance} & \textbf{Model} & \textbf{\ndcgau} & \textbf{\ndcga} & \textbf{\ndcgu}\\ 
		\midrule
		\multirow{9}{*}{$\geq$ 2.00}
		&\moqa        & 0.294 & 0.309  & 0.279\\
		&\moqatuning        & \textbf{0.319} & 0.212  & 0.480\\
		&\moqacp      & 0.318 & 0.212  & 0.477\\ \cline{2-5}
		
		&\filter      & 0.372 & 0.495  & 0.279\\
		&\filtertuning      & \textbf{0.516} & 0.400  & 0.666\\
		&\filtercp    & 0.514 & 0.392  & 0.675\\ \cline{2-5}
		
		&\bert        & 0.360 & 0.464  & 0.279\\
		&\berttuning        & 0.436 & 0.356  & 0.534\\
		&\bertcp      & \textbf{0.447} & 0.345  & 0.580\\ \midrule 

		\multirow{9}{*}{$\geq$ 2.25}
		&\moqa        & 0.264 & 0.249  & 0.279\\
		&\moqatuning        & \textbf{0.296} & 0.179  & 0.489\\
		&\moqacp      & 0.295 & 0.163  & 0.535\\ \cline{2-5}
		
		&\filter      & 0.361 & 0.468  & 0.279\\
		&\filtertuning      & 0.452 & 0.335  & 0.608\\
		&\filtercp    & \textbf{0.482}& 0.329  & 0.705\\ \cline{2-5}
		
		&\bert        & 0.344 & 0.423  & 0.279\\
		&\berttuning        & 0.373 & 0.293  & 0.477\\
		&\bertcp      & \textbf{0.405} & 0.310  & 0.530\\ \midrule 

		\multirow{9}{*}{$\geq$ 2.50}
		&\moqa        & 0.243 & 0.211  & 0.279\\
		&\moqatuning        & \textbf{0.274} & 0.165  & 0.453\\
		&\moqacp      & 0.265 & 0.155  & 0.452\\ \cline{2-5}
		
		&\filter      & 0.359 & 0.462  & 0.279\\
		&\filtertuning      & 0.439 & 0.326  & 0.592\\
		&\filtercp    & \textbf{0.470}& 0.316  & 0.699\\ \cline{2-5}
		
		&\bert        & 0.340 & 0.414  & 0.279\\
		&\berttuning        & \textbf{0.404} & 0.308  & 0.530\\
		&\bertcp      & 0.387 & 0.294  & 0.510\\ \midrule 
		
		\multirow{9}{*}{$\geq$ 2.75}
		&\moqa        & \textbf{0.235} & 0.199  & 0.279\\
		&\moqatuning        & 0.229 & 0.129  & 0.407\\
		&\moqacp      & 0.213 & 0.107  & 0.423\\ \cline{2-5}
		
		&\filter      & 0.333 & 0.397  & 0.279\\
		&\filtertuning      & 0.409 & 0.272  & 0.615\\
		&\filtercp    & \textbf{0.416} & 0.299  & 0.577\\ \cline{2-5}
		
		&\bert        & 0.330 & 0.390  & 0.279\\
		&\berttuning        & 0.349 & 0.279  & 0.435\\
		&\bertcp      & \textbf{0.388} & 0.296  & 0.509\\
		\bottomrule
	\end{tabular}
\end{table}

\subsection{Results}

We present the full results in ~\tabref{results}, reporting NDCG' 
performances over 
4 relevance thresholds: 
2.00, 
2.25, 
2.50, and 
2.75.

We'll first focus on the combined performances (\ndcgau). In general,  
all models (\moqa, \filter and \bert) see an improvement compared to 
their vanilla model when we tune a threshold ($+$\tuning or $+$\conform) 
to truncate the returned review list, implying it's helpful to find a 
cut-off to discover a more concise set of reviews.
Comparing between the simple thresholding method ($+$\tuning) vs.\ 
conformal method ($+$\conform), we also see very encouraging results: 
for both \filter and \bert, $+$\conform is consistently better than 
$+$\tuning for most relevance thresholds, suggesting that $+$\conform is 
a better rejection model. For \moqa, however, $+$\tuning is marginally 
better than $+$\conform. We hypothesize this may be due to \moqa 
producing an arbitrary (non-probabilistic) score for review, and as 
such is less suitable for conformal predictors. Comparing between the 
3 PQA models, \filter consistently produces the best performance: across 
  most relevance thresholds, \filtercp maintains an NDCG' performance 
close to 0.5. 


Looking at the \ndcgu results, we notice all vanilla models produce the 
same performance (0.279). This is because there are no relevant reviews 
for these unanswerable questions, and so the top-10 returned reviews by 
any models are always irrelevant. When we introduce $+$\tuning or 
$+$\conform to truncate the reviews, we see a very substantial 
improvement (\ndcgu more than doubled in most cases) for all models over 
different relevance thresholds. Generally, we also find that $+$\conform 
outperforms $+$\tuning, demonstrating that the conformal predictors are 
particularly effective for the unanswerable questions.


That said, when we look at the \ndcga performance, they are consistently 
worse when we introduce rejection ($+$\tuning or $+$\conform).  This is 
unsurprising, as ultimately it is a trade-off between precision and 
recall: when we introduce a rejection model to truncate the review list, 
we may produce a more concise/shorter list (as we see for the 
unanswerable questions), but we could also inadvertently exclude some 
potentially relevant reviews. As such, the best system is one that can 
maintain a good balance between pruning unreliable reviews and avoiding 
discarding potentially relevant reviews.


\begin{table}[t]
	
	\begin{center}	\caption{Reviews produced by \filter, \filtertuning 
			and \filtercp for an answerable (Q1) and unanswerable (Q2) 
			question.}
		\begin{tabular}{ p{1cm}  p{10cm}}
			\toprule
			\multicolumn{2}{l}{Q1: How long the battery lasts on X1 carbon touch?} \\
			\raisebox{-.5\totalheight}{\includegraphics[width=0.08\textwidth, height=8mm]{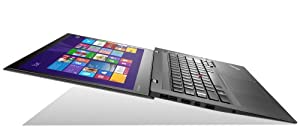}}
			& 
			\begin{tabular}{ p{4.cm}  p{6.cm}} 
				Ground Truth & [3, 3] \\
				\filter & [0, 0, 0, 0, 3, 0, 0, 0, 0, 0] \\
				\filtertuning & [0, 0, 0, 0, 3, 0, 0, 0] \\
				\filtercp & [0, 0, 0, 0, 3] \\
			\end{tabular}		
			\\ \midrule 
			
			\multicolumn{2}{l}{Q2: What type of memory SD card should I purchase to go with this? } \\
			\raisebox{-.5\totalheight}{\includegraphics[width=0.08\textwidth, height=8mm]{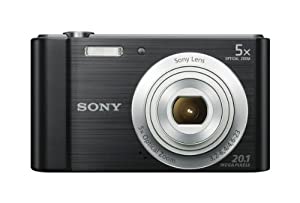}}
			& 
			\begin{tabular}{ p{4.cm}  p{6.cm}} 
				Ground Truth & [\phantom{0}] \\
				\filter & [0, 0, 0, 0, 0, 0, 0, 0, 0, 0] \\
				\filtertuning & [0, 0] \\
				\filtercp & [\phantom{0}] \\
			\end{tabular}		
			\\ \bottomrule 
		\end{tabular}
		\label{table:casestudy01}
	\end{center}
	
\end{table}

Next, we present two real output of how these methods perform in 
\tabref{casestudy01}.  The first question (Q1) is an answerable 
question, and the ground truth contains two relevant reviews (numbers in 
the list are review relevance scores).
\filter returns 10 reviews, of which one is relevant.
\filtertuning rejects the last two reviews, and
\filtercp rejects three more reviews, producing a concise list of 5 
reviews (no relevant reviews were discarded in this case).  One may 
notice both \filtertuning and \filtercp reject predictions but do not 
modify the original ranking of the returned reviews by \filter.
$+$\tuning tunes a threshold based on original relevance score, and in 
$+$\conform the conversion of class probability to confidence score 
($p$-value) is a monotonic transformation, and as such the original 
order is preserved in both methods.
This also means that if the vanilla model misses a relevant review, the 
review will not be recovered by the rejection model, as we see here.

The second question (Q2) is an unanswerable question (ground truth is an 
empty list).
\filter always returns 10 reviews, and so there are 10 irrelevant 
reviews.
\filtertuning discards most of the reviews, but there are still two 
irrelevant reviews.
\filtercp returns an empty list, indicating existing reviews do not have 
useful information to answer Q2, detecting correctly that Q2 is an 
unanswerable question.

\section{Conclusion}

PQA is often formulated as a retrieval problem with the goal to find the 
most relevant reviews to answer a given product question. In this paper,  
we propose incorporating conformal predictors as a rejection model to a 
PQA model to reject unreliable reviews. We test 3 state-of-the-art PQA 
models, \moqa, \filter and \bert, and found that incorporating conformal 
predictors as the rejection model helps filter unreliable reviews better 
than a baseline approach.  More generally, our paper demonstrates a 
novel and effective application of conformal predictors to a retrieval 
task.

\section*{Acknowledgement}
Shiwei Zhang is supported by the RMIT University and CSIRO Data61 Scholarships.


%
%
%


\end{document}